\documentclass[conference]{IEEEtran}
\IEEEoverridecommandlockouts
\usepackage{cite}
\usepackage{amsmath,amssymb,amsfonts}
\usepackage{pdfpages}
\usepackage{algorithmic}
\usepackage{graphicx} 
\usepackage{textcomp}
\usepackage{xcolor}
\usepackage{hyperref}    
\hypersetup{
    colorlinks=true, 
    linkcolor=blue, 
    citecolor=green, 
    urlcolor=magenta, 
    bookmarksopen=true 
}

\def\BibTeX{{\rm B\kern-.05em{\sc i\kern-.025em b}\kern-.08em
    T\kern-.1667em\lower.7ex\hbox{E}\kern-.125emX}}
\begin{document}

\title{Semantic-Guided Unsupervised Video Summarization\\
\thanks{Corresponding author: H. Yu, Email: hui.yu@glasgow.ac.u.k. This work was funded by UKRI (EP/Z000025/1) and the Horizon Europe Programme under the MSCA grant for ACMod project (Grant No. 101130271).}
}

\author{
\IEEEauthorblockN{1\textsuperscript{st} Haizhou Liu}
\IEEEauthorblockA{\textit{Department of Control Science and Engineering} \\
\textit{University of Shanghai for Science and Technology}\\
Shanghai 200093, China \\
233370837@st.usst.edu.cn}
\vspace{1.2em}   
\IEEEauthorblockN{3\textsuperscript{rd} Yiming Wang}
\IEEEauthorblockA{\textit{School of Psychology and Neuroscience} \\
\textit{University of Glasgow}\\
G12 8QQ Glasgow, U.K. \\
yiming.wang@glasgow.ac.uk}

\and

\IEEEauthorblockN{2\textsuperscript{nd} Haodong Jin}
\IEEEauthorblockA{\textit{Department of Control Science and Engineering} \\
\textit{University of Shanghai for Science and Technology}\\
Shanghai 200093, China\\
231260086@st.usst.edu.cn}
\vspace{1.2em}   
\IEEEauthorblockN{4\textsuperscript{th} Hui Yu}
\IEEEauthorblockA{\textit{School of Psychology and Neuroscience} \\
\textit{University of Glasgow}\\
G12 8QQ Glasgow, U.K. \\
hui.yu@glasgow.ac.uk}
}
\maketitle

\begin{abstract}
Video summarization is a crucial technique for social understanding, enabling efficient browsing of massive multimedia content and extraction of key information from social platforms. Most existing unsupervised summarization methods rely on Generative Adversarial Networks (GANs) to enhance keyframe selection and generate coherent, video summaries through adversarial training. However, such approaches primarily exploit unimodal features, overlooking the guiding role of semantic information in keyframe selection, and often suffer from unstable training. To address these limitations, we propose a novel Semantic-Guided Unsupervised Video Summarization method. Specifically, we design a novel frame-level semantic alignment attention mechanism and integrate it into a keyframe selector, which guides the Transformer-based generator within the adversarial framework to better reconstruct videos. In addition, we adopt an incremental training strategy to progressively update the model components, effectively mitigating the instability of GAN training. Experimental results demonstrate that our approach achieves superior performance on multiple benchmark datasets.
\end{abstract}

\begin{IEEEkeywords}
Semantic Guidance, Social Intelligence, Video Summarization.
\end{IEEEkeywords}

\section{Introduction}
With the rapid proliferation of cameras and recording devices, massive amounts of video data are captured every day, documenting a wide range of social events. However, the sheer volume of video content poses a major challenge: how to efficiently search, classify, and browse specific information of interest. To address this issue, video summarization~\cite{11113316},~\cite{10982110},~\cite{11130654} has emerged as an effective technique that enables users to quickly browse large-scale video collections and extract key information. It has been widely applied in diverse social-visual scenarios such as event highlight detection~\cite{li2024unsupervised},~\cite{wang2024modality}, traffic prediction~\cite{10855347},~\cite{10706115}, disaster monitoring~\cite{shi2022application},~\cite{qin2024sora}, and security surveillance~\cite{goyal2023captionomaly},~\cite{10287367}.

Driven by the advancement of deep learning, three categories of video summarization methods have been developed: supervised~\cite{zhu2022relational}, weakly supervised~\cite{11086416}, and unsupervised approaches~\cite{11077719}. Supervised methods rely on large-scale paired annotations to guide the model in identifying salient content. However, the manual annotation process is costly, time-consuming, and inherently subjective. Weakly supervised methods reduce annotation cost by introducing prior knowledge or partial labels to improve generalization, yet they still suffer from label bias and difficulties in capturing the complexity and diversity of video semantics. In contrast, unsupervised methods have attracted increasing attention because they do not require human labels. Instead, they generate summaries through heuristic strategies, with adversarial learning~\cite{10520276} and reinforcement learning~\cite{9690701} being the most widely used approaches.

Reinforcement learning methods improve representativeness and diversity by designing reward functions, but their effectiveness heavily depends on the reward design and often struggles to capture long-range frame dependencies. Generative Adversarial Networks (GANs), on the other hand, employ adversarial learning to drive sequence generation models for unsupervised video summarization. Existing GAN-based methods typically adopt recurrent neural networks (e.g., LSTMs)~\cite{zhang2020unsupervised},~\cite{8765236},~\cite{10947580} as generators to capture multi-scale temporal dependencies. Although recurrent networks can model contextual relations between frames, they are inefficient in handling long-term dependencies and parallel training. Moreover, most current approaches focus solely on unimodal feature reconstruction, neglecting the inherent visual–semantic correlations in videos, and thus fail to effectively model multimodal interactions.

For a complete video, its underlying content is uniquely determined and remains fixed. Consequently, different modalities of the same video (e.g., visual and semantic information) inherently convey the same thematic content. In other words, strong correlations naturally exist across modalities within a video. Specifically, the visual and textual features of the same frame should exhibit a deterministic mapping, and therefore the data distributions between the original video and its cross-modal summaries should also align accordingly. To capture semantic information, we leverage the CLIP encoder~\cite{radford2021learning}, which possesses zero-shot classification capability, to extract latent semantic features from video frames. By computing the cosine similarity between the visual and semantic feature sequences, we establish cross-modal alignment and obtain a quantitative measure of their correspondence. This similarity score is then regarded as the importance score assigned by the frame selector to each frame, serving as guidance for summary generation. Inspired by this idea, we exploit semantic signals within video frame sequences as cues to drive adversarial training, enabling the model to generate summaries that maximally represent the original video content.

Hence, we propose a Semantic-Guided Unsupervised Video Summarization method. As illustrated in Fig.~\ref{fig:mainfig}, we first employ a CNN and a CLIP encoder to extract visual and textual features, respectively. A frame selector with integrated frame-level semantic alignment attention then computes keyframe scores to guide a Transformer-based generator in reconstructing the video summary. Training proceeds until the discriminator can no longer distinguish between real videos and generated summaries, thus achieving semantic-guided adversarial learning. In addition, we adopt an incremental training strategy to progressively update the adversarial components, effectively mitigating the instability of GAN training. 

The main contributions of this work are summarized as follows:
\begin{enumerate}
    \item We design a novel adversarial learning framework driven by frame-level semantic alignment attention, and introduce an incremental training strategy to effectively stabilize GAN training.
    \item We propose a new attention mechanism that quantifies the importance of keyframes as the cosine similarity across modalities, enabling cross-modal interaction and joint modeling of inherent visual–semantic correlations, thereby improving summary quality.
    \item We design a Transformer-based generator with a self-encoding architecture to capture both textual and visual dependencies, enabling adaptive learning of multimodal features to compensate for the limitations of unimodal modeling and produce high-quality summaries.
\end{enumerate}

\section{Related Work}

\subsection{Unsupervised Video Summarization}

Unsupervised video summarization methods~\cite{zang2024video} typically adopt heuristic strategies for video reconstruction, which can be broadly categorized into reinforcement learning RL-based and generative adversarial network GAN-based approaches. RL-based methods rely on carefully designed reward functions to guide keyframe selection. However, the effectiveness of such methods is highly dependent on reward design, which is often difficult and task-specific. GAN-based methods, in contrast, enable unsupervised learning through an adversarial training paradigm. A summary generator is trained to produce compact summaries, while a discriminator tries to distinguish these summaries from the original video. Over the course of training, the generator gradually improves, ultimately producing high-quality video summaries capable of fooling the discriminator. Mahasseni et al. ~\cite{mahasseni2017unsupervised} first introduced the adversarial paradigm into unsupervised video summarization by minimizing the distribution discrepancy between original videos and generated summaries, and thereby laying the foundation for this line of research. Building upon this work, Apostolidis et al. ~\cite{apostolidis2019stepwise} proposed the integration of a linear compression layer between the input and the summarizer to reduce feature dimensionality, along with a stepwise label-based learning strategy to improve adversarial training efficiency. Later, they further extended this approach by introducing the SUM-GAN-AAE model~\cite{apostolidis2020unsupervised}, which incorporated attention mechanisms into both the variational autoencoder and the autoencoder layers, significantly enhancing the model’s ability to capture contextual dependencies. Although GAN-based unsupervised video summarization~\cite{10029884},~\cite{9259058} has achieved notable progress, these methods are still constrained by the inherent instability of GAN training and related challenges, leaving considerable room for performance improvement in summary generation.

\begin{figure*}[!h]
    \centering
    \includegraphics[width=0.9\linewidth]{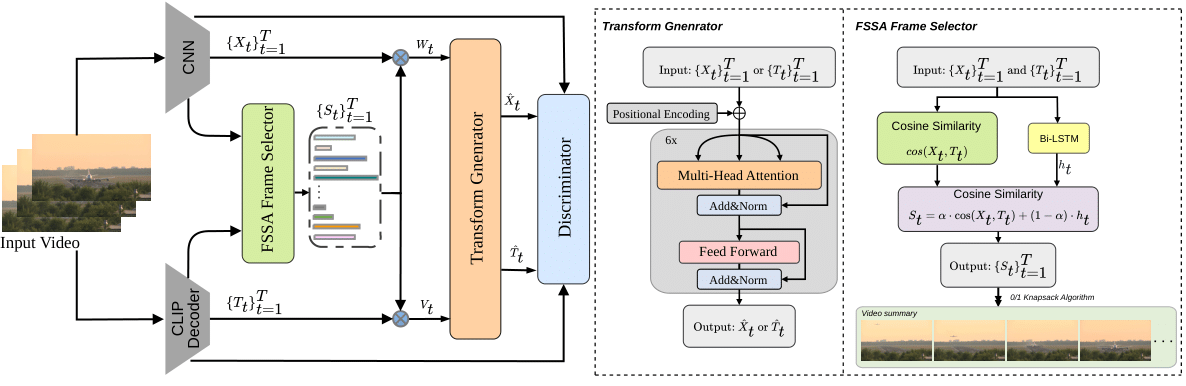}
    \caption{\textbf{ Overview workflow of the proposed framework for unsupervised video summarization.} 
         Symbols: $\oplus$ represents Concatenation, $\otimes$ represents Element-wise multi.}
    \label{fig:mainfig}
\end{figure*}

\section{Proposed Method}

\subsection{Overview of Technical Framework}

We formulate video summarization as a structured prediction problem: given a sequence of continuous video frames as input, the goal is to output a sequence of keyframes that effectively represents the original video. As illustrated in Fig.~\ref{fig:mainfig}, a video sequence containing $T$ frames is first processed by a CNN and a CLIP encoder to extract visual features $\{X_t\}_{t=1}^{T}$ and semantic features $\{T_t\}_{t=1}^{T}$ for each frame. Subsequently, the multimodal features of the frame sequence are fed into a frame selector, which produces a frame importance score sequence $\{S_t\}_{t=1}^{T}$. Each score $S_t$ lies within the range $[0,1]$, where a higher value indicates greater importance of frame $t$ in the video. Based on these importance scores, the visual and semantic features of each frame are re-weighted as $W_t = S_t \cdot X_t$ and $V_t = S_t \cdot T_t$, and then passed into a Transformer-based generator to produce the reconstructed summary features $\{\hat{X}_t\}$ and $\{\hat{T}_t\}$. Finally, the reconstructed multimodal summary features $(\hat{X}_t, \hat{T}_t)$ together with the original features $(X_t, T_t)$ are jointly fed into a discriminator, such that the adversarial training framework ensures the generated summaries closely approximate the semantic and visual distribution of the original video.  

\subsection{Frame Selector with Semantic-alignment Attention}

The frame selector assigns an importance score to each frame to measure its representativeness within the entire video. Based on these scores, the visual and semantic features of frames are reweighted to form a new sequence feature distribution. The more accurate the frame scores are, the better the sequence can preserve frames that maximally represent the original video content, thereby facilitating the generator to reconstruct higher-quality summaries. However, most existing methods adopt bidirectional LSTM (Bi-LSTM) networks~\cite{8931648} to capture temporal dependencies between frames and assign scores accordingly. Such approaches often rely solely on modeling visual features of video sequences while neglecting the contribution of semantic information to frame scoring.

To address this limitation, we propose a Frame-level Semantic-alignment Attention (FSSA) mechanism. This mechanism incorporates the cosine similarity between the visual and semantic features of video frames into the frame score sequence, allowing semantic information to serve as guidance during scoring and ultimately enabling the generated summaries to better capture the core content of the video.

Formally, the frame score is defined as
\begin{equation}
S_t = \alpha \cdot \text{cos}(X_t, T_t) + (1 - \alpha) \cdot h_t
\end{equation}
where $S_t$ denotes the importance score of frame $t$, $\text{cos}(X_t, T_t)$ represents the cross-modal cosine similarity between the visual feature $X_t$ and the semantic feature $T_t$, and $h_t$ is the score produced by the Bi-LSTM. A higher $S_t$ indicates that the abstracted deep feature of the current frame is more representative of the overall video content.  

To fully exploit the advantage of cross-modal attention in frame selection, we introduce a learnable parameter $\alpha \in [0,1]$. During training, $\alpha$ dynamically adjusts the relative weights between the cross-modal similarity score and the Bi-LSTM score, enabling the model to adaptively determine the optimal scoring scheme for different video contexts.

\subsection{Unsupervised Learning Strategy}

For a video containing $T$ frames, we denote the original visual and semantic feature vectors of the sequence as $\{X_t, T_t\}_{t=1}^{T}$, and the reconstructed features as $\{\hat{X}_t, \hat{T}_t\}_{t=1}^{T}$.  

The training procedure is carried out in three stages. First, a forward pass is performed to compute the reconstruction losses between $(X_t, T_t)$ and $(\hat{X}_t, \hat{T}_t)$, and the discriminator is updated accordingly. Second, with partially updated parameters, another forward pass is performed to compute the discrepancy between $T_t$ and $\hat{T}_t$, after which the frame selector is updated. Finally, a third forward pass is conducted, where the outputs $\hat{X}_t$ are further compared against $X_t$, and both the frame selector and generator are optimized to minimize the reconstruction loss.  

The overall unsupervised training objective combines reconstruction and sparsity losses. Specifically, the reconstruction loss is defined as  
\begin{equation}
L_{\text{rec}} = \mu \cdot \|X_t - \hat{X}_t\|_2^2 + 
\nu \cdot \|T_t - \hat{T}_t\|_2^2 + \text{SSIM}(X_t, \hat{X}_t)
\end{equation}
where $X_t$ and $T_t$ denote the original video features, $\hat{X}_t$ and $\hat{T}_t$ are the reconstructed features, 
$\mu$ and $\nu$ are hyperparameters, and $\text{SSIM}$ denotes the structural similarity loss~\cite{7104105}.  

To enforce compactness and avoid trivial solutions, we further introduce a sparsity constraint on the frame importance scores:  
\begin{equation}
L_{\text{sparsity}} = \left| \frac{1}{N} \sum_{j=1}^N \big(S_t^j - \lambda\big) \right|
\end{equation}
where $S_t^j$ represents the importance score of the $j$-th frame in the sequence, $\lambda$ is a sparsity regularization hyperparameter, and $N$ is the total number of frames.  

The overall training loss integrates both reconstruction and sparsity constraints, and is defined as
\begin{equation}
L_{\text{total}} = L_{\text{rec}} + L_{\text{sparsity}}
\end{equation}
which jointly ensures accurate feature reconstruction while promoting compact and informative video summaries.

\section{Experiments}

\subsection{Datasets and Experimental Settings}

We evaluate the proposed semantic-guided unsupervised video summarization model on two widely used benchmark datasets: SumMe and TVSum.  

SumMe~\cite{gygli2014creating} contains 25 videos ranging from 1 to 6 minutes in length, covering multi-view social events. Each video is annotated by 15–18 users, who generated summaries by selecting key shots, and frame-level importance scores are provided as a single reference ground truth.  

TVSum~\cite{song2015tvsum} consists of 50 videos with durations from 1 to 11 minutes. Each video is annotated by 20 users, who assigned frame-level importance scores on a scale of 1–5. The summary length is constrained to 5\%–15\% of the total video duration.  

Performance is evaluated using the F1-score, computed as the harmonic mean of precision ($P$) and recall ($R$):  
\begin{equation}
F1 = \frac{2PR}{P+R}
\end{equation}
where the overlap between the automatically generated summary ($A$) and the user-generated summary ($G$) is measured at the frame level. For TVSum, the mean F1-score across all users is reported, while for SumMe, the maximum F1 score among annotators is used. To ensure summary length constraints, we adopt the backpack algorithm to restrict summaries within 15\% of the video duration.  

For experimental settings, all videos are downsampled to 2 fps. The model is implemented in PyTorch and trained on an NVIDIA GeForce RTX 3090 GPU. We use the Adam optimizer with a learning rate of $\lambda = 0.001$ for the discriminator and $\lambda' = 0.002$ for the other modules. Each dataset is randomly split into 80\% for training and 20\% for testing, and the reported results are averaged over five independent runs.  

\subsection{Quantitative Comparison}

To ensure evaluation fairness, we adopt dual assessment frameworks for quantifying algorithm-generated summary quality: one framework leverages multi-annotator summaries per video, whereas the other employs a single reference summary. As shown in Table~\ref{tab:comparison1} and Table~\ref{tab:comparison2}, our method achieves the best performance on both the SumMe and TVSum datasets. In addition, we have observed that the F1-scores obtained under the single-reference evaluation setting are consistently higher than those under the multi-user annotation setting. One possible reason is that the evaluation in Table~\ref{tab:comparison1}, which aggregates multiple user annotations, provides a more objective benchmark. In contrast, the evaluation in Table~\ref{tab:comparison2}, which relies on a single reference summary, may be affected by the subjectivity of the annotator. These results highlight the necessity of adopting diverse evaluation metrics, as they capture different aspects of summarization performance and ensure a more comprehensive assessment of model effectiveness. 

Generally, our method achieves the highest average F1 score on both benchmark datasets, outperforming the second-best result by 3.2 points. This consistent improvement demonstrates the effectiveness of our approach in capturing key video content. In addition, the results show that the F1 scores on the TVSum dataset are consistently higher than those on SumMe. A likely explanation is that TVSum contains a larger number of videos with longer durations and more diverse content. These characteristics provide richer semantic information that allows our model to learn better and more generalizable features.

\begin{table}[htbp]
    \centering
    \caption{The summarization approach using multiple user annotations is compared with other unsupervised methods. (“±” indicates better or worse performance than our method).}   
    \begin{tabular}{lccc}
            \hline
            Methods & SumMe & TVSum & Average \\
            \hline
            Tessellation & 41.4(-) &64.1(-) & 52.8(-) \\
            DR-DSN & 41.4(-) & 57.6(-) & 49.5(-) \\
            DSR-RL-LSTM & 43.8(-) & 61.4(-) & 52.6(-) \\
            DSR-RL-GRU & 50.3(-) & 60.2(-) & 55.3(-) \\
            UnpairedVSN & 47.5(-) & 55.6(-) & 51.6(-) \\
            SUM-Ind\_LU & 51.9(-) & 61.5(-) & 56.7(-) \\
            AC-SUM-GAN & 50.8(-) & 60.6(-) & 55.7(-) \\
            CAAN & 50.8(-) & 59.6(-) & 55.2(-) \\
            CSNet & 51.3(-) & 58.8(-) & 55.1(-) \\
            Cycle-SUM & 41.9(-) & 57.6(-) & 49.8(-) \\
            RS-SUM & 52.0(-) & 61.1(-) & 56.6(-) \\
            SUM-GDA & 50.0(-) & 59.6(-) & 54.8(-) \\
            SUM-GAN-sl & 47.3(-) & 58.0(-) & 52.7(-) \\
            SUM-GAN-AAE & 48.9(-) & 58.3(-) & 53.6(-) \\
            SUM-GAN-GEA & 53.4(-) & 61.3(-) & 57.4(-) \\
            DMFF & 51.8(-) & 61.0(-) & 56.4(-) \\
            PRLVS & 46.3(-) & 63.0(-) & 54.7(-) \\
            SSPVS & 48.7(-) & 60.3(-) & 54.5(-) \\
            M3Sum(SP) & 43.6(-) & 56.9(-) & 50.2(-) \\
            AMFM & 51.8(-) & 61.0(-) & 56.4(-) \\
            Ours  &\textbf{55.9} & \textbf{65.3} &\textbf{60.6} \\
            \hline
    \end{tabular}
    \label{tab:comparison1}
\end{table}

\begin{table}[htbp]
    \centering
    \caption{Comparison with other video summarization methods using single ground truth summarization (asterisks indicate unsupervised methods, “$\pm$” indicate better or worse performance than our method).}
    \begin{tabular}{lcc}
        \hline
        Methods & SumMe & TVSum \\
        \hline
        SASUM & 45.3(-) & 58.2(-) \\
        DTR - GAN & 44.6(-) & 59.1(-) \\
        A - AVS & 43.9(-) & 59.4(-) \\
        M - AVS & 44.4(-) & 61.0(-) \\
        AALVS & 46.2(-) & 63.6(-) \\
        *DMFF & 62.7(-) & 65.3(-) \\
        *Cycle - sum & 41.9(-) & 57.6(-) \\
        *SUM - GAN - sl & 46.8(-) & 65.3(-) \\
        *SUM - GAN - AAE & 56.9(-) & 63.9(-) \\
        *SUM - GAN - GEA & 65.9(-) & 68.2(-) \\
        *AC - SUM - GAN & 60.7(-) & 64.8(-) \\
        *SUM - GAN & 38.7(-) & 50.8(-) \\ 
        *SUM - GAN$_{\text{dpp}}$ & 39.1(-) & 51.7(-) \\
        *SUM - GAN$_{\text{sup}}$ & 41.7(-) & 56.3(-) \\
        *Cycle - SUM$_{\text{dpp}}$ & 42.3(-) & 57.9(-) \\
        *Cycle - SUM$_{\text{sup}}$ & 44.8(-) & 58.1(-) \\
        *Ours &\textbf{63.7} &\textbf{70.1} \\
        \hline
    \end{tabular}
    \label{tab:comparison2}
\end{table}    

\subsection{Qualitative Comparison}

\begin{figure}[!t]
    \centering 
    \includegraphics[width=0.5\textwidth]{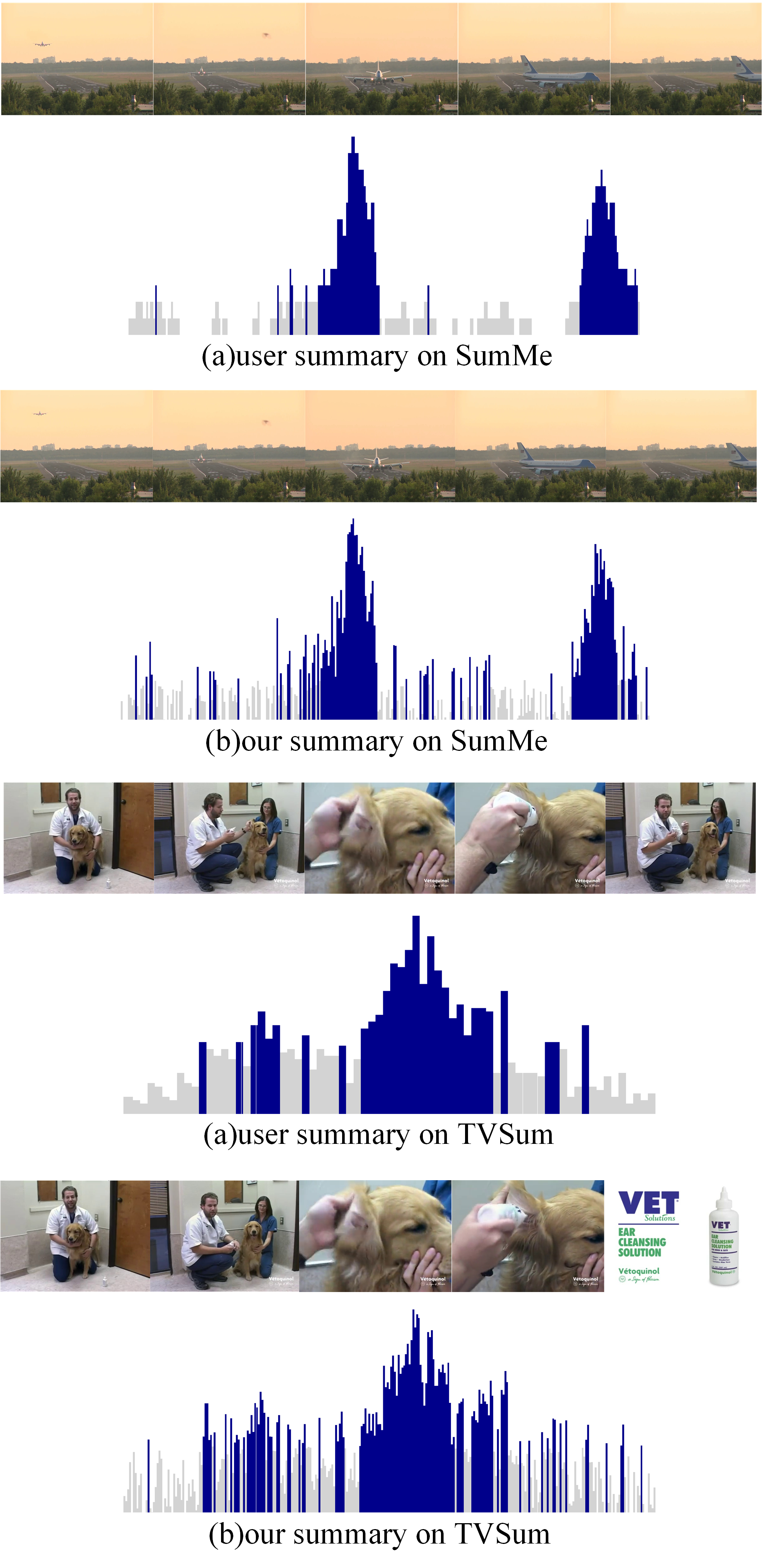} 
    \caption{ Partial summary visualization results generated by the proposed method. Gray is ground truth, blue is the summary result of model prediction, and the height of the column indicates the importance of the frame.The video summary on the left is from the first video in the SumMe dataset, while the one on the right is from the 15th video in the TVSum dataset.}
    \label{fig:videosummary_show} 
\end{figure}

Figure~\ref{fig:videosummary_show} presents visual examples of the summaries generated by our model on the two benchmark datasets. In the SumMe dataset, the first example depicts an airplane landing, while in the TVSum dataset, the 15th example demonstrates the use of a product to treat a dog’s ears. These representative cases highlight the diversity of video content across datasets, providing a meaningful basis for evaluating summary quality. Part (a) of the figure shows the ground-truth summaries annotated by human subjects, together with histograms of keyframe score distributions. Part (b) illustrates the summaries generated by our model and the corresponding predicted score distributions. Comparing the two reveals that the majority of keyframes selected by the model align closely with the peaks observed in the ground-truth distributions. This alignment indicates that our method not only identifies the most salient moments but also captures the semantic continuity of events. As a result, the generated summaries reflect human subjective perception more faithfully, demonstrating that the model balances informativeness with coherence.

\subsection{Ablation Study}

Table~\ref{tab:ablation_study} reports the performance of different model variants on both the SumMe and TVSum datasets. The results clearly demonstrate that the proposed method significantly outperforms traditional unsupervised video summarization approaches based on Generative Adversarial Networks (GANs). From the comparison, it can be observed that the FSSA frame selector effectively facilitates cross-modal information interaction. By leveraging the frame-level attention distribution derived from semantic–visual alignment, it guides the reconstruction of more representative and higher-quality summaries. In addition, the Transformer-based generator further enhances cross-modal perception and enables fine-grained summary reconstruction, which is consistently reflected in the improvements reported in Table~\ref{tab:ablation_study}. The results of Exp5 appear less promising, showing performance comparable to that of using a single strategy. However, this outcome provides strong evidence of the importance of semantic guidance in our approach. Specifically, the Frame Selector with Semantic-Alignment Attention serves as the core of our unsupervised framework and plays a decisive role in effective video reconstruction.

\begin{table}[htbp]
    \centering
    \caption{Ablation study.}
    \resizebox{\columnwidth}{!}{ 
        \begin{tabular}{lcccccc}
            \hline
            Experiments & Multimodal & FSCA & Transform Gnenrator & Summe & TvSum \\
            \hline
            SUM - GAN - AAE &  &  &  & 48.9 & 58.3 \\
            Exp1 & $\checkmark$ &  &  & 49.3 & 59.6 \\
            Exp2 &  & $\checkmark$ &  & 52.7 & 62.4 \\
            Exp3 &  &  & $\checkmark$ & 51.5 & 60.3 \\
            Exp4 & $\checkmark$ & $\checkmark$ &  & 52.4 & 63.6 \\
            Exp5 & $\checkmark$ &  & $\checkmark$ & 51.9 & 59.1 \\
            Exp6 &  & $\checkmark$ & $\checkmark$ & 53.9 & 64.3 \\
            Ours & $\checkmark$ & $\checkmark$ & $\checkmark$ & 55.9 & 65.3 \\
            \hline
        \end{tabular}
    }
    \label{tab:ablation_study}
\end{table}

\section{Conclusion and Future Work}

In this paper, we proposed a novel unsupervised video summarization framework, termed Semantic-Guided Unsupervised Video Summarization. The core idea of our approach is to compute cross-modal cosine similarity between frame-level representations within the same video sequence and use this similarity distribution as semantic cues to guide summary generation. The entire framework is trained with an incremental optimization strategy, ensuring stable and effective adversarial learning. Extensive experiments on benchmark datasets demonstrate the effectiveness and superiority of the proposed method over existing approaches.  

In future work, we plan to incorporate audio features into the framework to further enhance summary quality. By exploring richer multimodal feature representations, we aim to extend the applicability of semantic-guided video summarization to more diverse real-world scenarios.

\bibliographystyle{IEEEtran}  
\bibliography{ref}      

\end{document}